\documentclass[10pt,twocolumn,letterpaper]{article}

\usepackage{icb}
\usepackage{times}
\usepackage{epsfig}
\usepackage{graphicx}
\usepackage{amsmath}
\usepackage{amssymb}
\usepackage[caption=false,font=footnotesize]{subfig}
\usepackage[font={footnotesize}]{caption}
\usepackage[table]{xcolor}
\usepackage{tabularx}
\usepackage[flushleft]{threeparttable} 
\usepackage{booktabs,caption}
\usepackage{fancyhdr}

\usepackage{hyperref}
\hypersetup{
  colorlinks, linkcolor=red
}
\newcolumntype{Y}{>{\centering\arraybackslash}X}
\newcolumntype{s}{>{\hsize=.3\hsize}X}
\newcolumntype{b}{>{\hsize=1.4\hsize}X}
\makeatletter
\g@addto@macro{\UrlBreaks}{\UrlOrds}
\makeatother
\makeatletter
\renewcommand\subsubsection{\@startsection{subsubsection}{3}{\z@}%
                       {-8\p@ \@plus -4\p@ \@minus -4\p@}
                       {-0.5em \@plus -0.22em \@minus -0.1em}%
                       {\normalfont\normalsize\bfseries\boldmath}}
\makeatother

\makeatletter
\def\and{%
  \end{tabular}%
  \hskip 0.1em \@plus.17fil\relax
  \begin{tabular}[t]{c}}
\makeatother

\icbfinalcopy 

\ificbfinal\fi
\begin{document}


\title{\vspace{-1em}Longitudinal Study of Child Face Recognition}

\author{Debayan Deb\\
Michigan State University\\
East Lansing, MI, USA\\
{\tt\small debdebay@msu.edu}
\and
Neeta Nain\\
Malaviya National Institute of Technology\\
Jaipur, India\\
{\tt\small nnain.cse@mnit.ac.in}
\and
Anil K. Jain\\
Michigan State University\\
East Lansing, MI, USA\\
{\tt\small jain@cse.msu.edu}
}

\maketitle
\thispagestyle{empty}

\begin{abstract}
We present a longitudinal study of face recognition performance on Children Longitudinal Face (CLF) dataset containing $3,682$ face images of $919$ subjects, in the age group $[2, 18]$ years. Each subject has at least four face images acquired over a time span of up to six years. Face comparison scores are obtained from (i) a state-of-the-art COTS matcher (COTS-A), (ii) an open-source matcher (FaceNet), and (iii) a simple sum fusion of scores obtained from COTS-A and FaceNet matchers. To improve the performance of the open-source FaceNet matcher for child face recognition, we were able to fine-tune it on an independent training set of 3,294 face images of 1,119 children in the age group $[3, 18]$ years. Multilevel statistical models are fit to genuine comparison scores from the CLF dataset to determine the decrease in face recognition accuracy over time. Additionally, we analyze both the verification and open-set identification accuracies in order to evaluate state-of-the-art face recognition technology for tracing and identifying children lost at a young age as victims of child trafficking or abduction. 
\end{abstract}

\section{Introduction}

The United Nations Convention on the \emph{Rights of the Child} defines child as ``a human being below the age of $18$ years unless under the law applicable to the child, majority is attained earlier''~\cite{child-definition}. This definition is ratified by $192$ of the $194$ countries that are members of the United Nations. According to the United Nations Children's Fund (UNICEF), nearly 2 million children under the age of 20 are subjected to prostitution in the global sex trade. On average, victims range from 11 to 14 years old and are expected to survive only 7 years. The United Nations Office on Drugs and Crime reports the percentage of child trafficking victims has risen about $25$\% from $2009$ to $2012$, where the victims are in the age group of $1$ to $18$ years~\cite{unodc}. For every three child victims, two are girls and one is a boy.
%
According to Kolkata's \emph{Child in Need Institute}, $1,628$ kidnapped children, in the age group of $4$ to $15$ years, were retrieved from a single railway station; among these, $134$ were girls and the youngest was only four years old~\cite{guardian-report}. Of course, these are official statistics, and do not necessarily reflect the true numbers of child kidnapping and sex trafficking in a population of around 1.2 billion in India.

\begin{table*}[!t]
\footnotesize
\caption{Related work on longitudinal study of face recognition.}
\centering
\begin{threeparttable}
\renewcommand{\arraystretch}{1.5}
\begin{tabularx}{\textwidth}{>{\centering\bfseries}l>{}X >{\centering}l >{\arraybackslash}X}
\noalign{\hrule height 1.5pt}
Study & Objective & Dataset & Findings \\
   \noalign{\hrule height 1pt}
 Otto~\etal~\cite{otto-jain}\tnote{*} & Influence of facial aging on different facial components. & MORPH-II & The nose is the most stable component across face aging.\\
   \hline
   Bereta~\etal~\cite{bereta}\tnote{*} & Investigation of local descriptors for face recognition in the context of age progression. & FG-NET & Accuracy for local descriptors combined with Gabor magnitudes are most stable.\\
    \hline
   Ricanek~\etal~\cite{ricanek} & Face aging effects on face recognition (from infant to adulthood). & ITWCC & 24\% TAR at 0.1\% FAR for verification scenario. Rank-1 identification performance is 25\%.\\
    \hline
    Deb~\etal~\cite{Deb} & Analysis of rates of change in genuine scores over time due to facial aging. & PCSO, MSP & COTS matchers can verify 99\% of the subjects at a FAR of 0.01\% for up to 10.5 years of elapsed time.\\
    \hline
    Best-Rowden~\etal~\cite{Lacey1} & Investigate the feasibility of automatic face recognition for children in the age group of $0$ to $4$ years. & NITL & 47.93\% TAR at 0.1\% FAR ($\Delta T = 6$ months).\\
    \hline 
    Basak~\etal~\cite{iitd} & Evaluation of multimodal biometric recognition for children in the age group of $2$ to $4$ years. & CMBD  & 19\% TAR at 0.1\% FAR for a single face image/subject in the gallery.\\
    \hline
    \rowcolor{gray!50}This study & Investigate the feasibility of automatic face recognition for children in the age group of $2$ to $18$ years. & CLF & 90.18\% TAR at 0.1\% FAR  ($\Delta T = 1$ year).\\
\noalign{\hrule height 1.5pt}
\end{tabularx}
\begin{tablenotes}\footnotesize
\item[] TAR = true accept rate; FAR = false accept rate; $\Delta T$ = time lapse between enrollment and probe image
\item[*] This study is considered cross-sectional study and not longitudinal as age group is partitioned into smaller ranges~\cite{Lacey1},~\cite{Yoon},~\cite{msm}
\end{tablenotes}
 \end{threeparttable}
\label{tab:related}
\end{table*}

\begin{table*}[!t]
\footnotesize
\caption{Table of longitudinal face datasets.}
\centering
\begin{tabularx}{0.885\linewidth}{l l l l l l l l}	
\noalign{\hrule height 1.5pt}
Dataset & No. of Subjects & No. of Images & No. Images / Subject & Age Group (years) & Avg. Age (years)  & Public\protect\footnotemark\\
   \noalign{\hrule height 1pt}
  MORPH-II~\cite{Morph} &13,000 & 55,134 & 2-53 (avg. 4) &16-77 & 42 & Yes\\
   \hline
   FG-NET~\cite{fgnet} & 82 & 1,002 & 6-18 (avg. 12) & 0-69 & 16 &  Yes\\
    \hline
    ITWCC~\cite{ricanek} & 304 & 1,705 & 3+ & 5 mos. - 32 yrs & 13 & No\\
    \hline
    PCSO~\cite{Deb} & 18,007 & 147,784  & 5-60 (avg. 8) & 18-83 & 31 & No\\
    \hline
     MSP~\cite{Deb} & 9,572 & 82,450 & 4-48 (avg. 9) & 18-78 & 33 & No\\
    \hline
     NITL~\cite{Lacey1} & 314 & 3,144 & 3-5 & 0-4 & N/A & No\\
    \hline
    CMBD~\cite{iitd} & 106 & 1,060 & 10 & 2-4 & N/A & No\\
    \hline
    \rowcolor{gray!50}CLF (this study) & 919 & 3,682 & 2-6 (avg. 4) & 2-18 & 8 & No\\
\noalign{\hrule height 1.5pt}
\end{tabularx}
\label{tab:allDatasets}
\end{table*}

\begin{figure}[!t]
\centering
\subfloat[]{\includegraphics[height=1.5in,keepaspectratio]{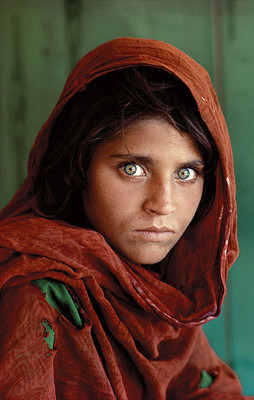}
\label{fig:gula1}}\hfil
\subfloat[]{\includegraphics[height=1.5in,keepaspectratio]{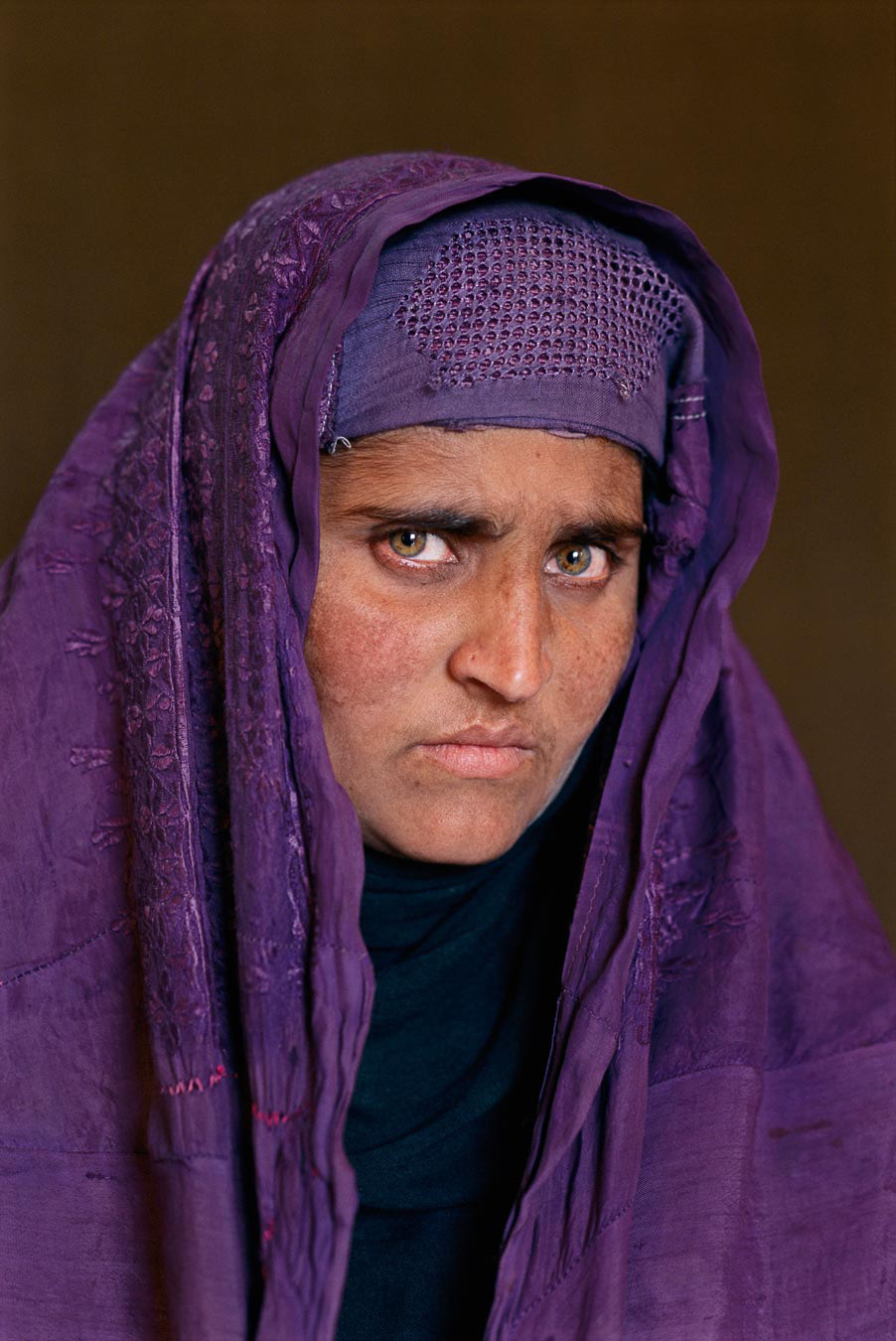}
\label{fig:gula2}}
 \caption{Sharbat Gula (a) at age 12, photographed in 1984 and (b) at age 30, photographed in 2002~\cite{nat-geo}. She was identified based on iris recognition~\cite{daugman}.}
\label{fig:gula}
\end{figure}

\begin{figure}[!t]
\centering

\subfloat[]{\resizebox{!}{0.9in}{\includegraphics[width=\linewidth,keepaspectratio]{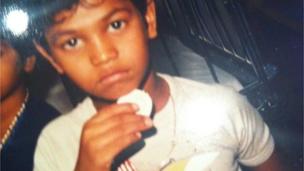}}}
\hfil
\subfloat[]{\resizebox{!}{0.9in}{\includegraphics[width=\linewidth,keepaspectratio]{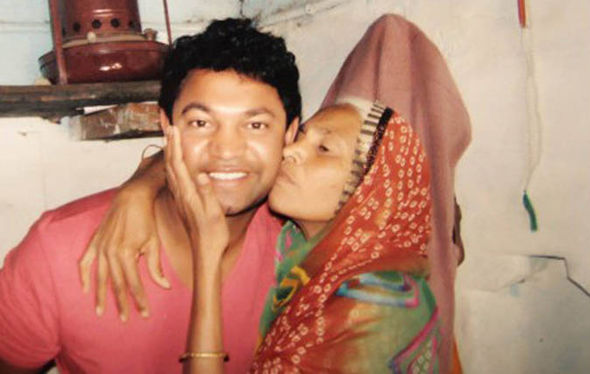}}}
 \caption{Saroo Breirley (a) before he went missing at age 5 and (b) after reuniting with his biological mother at age 30~\cite{a-long-way-home}.}
\label{fig:saroo}
\end{figure}


To trace missing children, face recognition is perhaps the primary biometric modality since parents and relatives are more likely to have a lost child's photographs(s) as opposed to, say, fingerprint or iris. However, face recognition is certainly not the only biometric modality for identification of lost children. Sherbat Gula, first photographed by the photographer Steve McCurry in 1984 (age 12)  in a refugee camp in Pakistan (Figure~\ref{fig:gula1}), was traced at the age of 30 to a remote part of Afghanistan where she was photographed again (Figure~\ref{fig:gula2}) in 2002~\cite{nat-geo}. Daugman magnified the eye regions in both the 1984 and 2002 photographs and confirmed Gula's identity using iris recognition~\cite{daugman}. 

\footnotetext{{\label{datasetLinks}MORPH-II is available at \url{https://ebill.uncw.edu/C20231_ustores/web/store_main.jsp?STOREID=4} and FG-NET is available at \url{http://yanweifu.github.io/FG_NET_data/index.html}.}}
It is often the case that there is a huge gap (in years) between the time a child is lost and retrieved. For example, Saroo Brierley\footnote{The award-winning 2016 movie, \emph{Lion}, is based on the true story of Saroo Brierley~\cite{lion-movie}.}  (also known as Sheru, which stands for  ``lion'' in Hindi) was lost at the age of $5$ from Khandwa railway station in India, and later adopted by Australian parents, Sue and John Brierley. Saroo was reunited with his family as an adult, at the age of $30$; his biological mother could identify him through his $5$ year old pictures maintained by the Brierleys. Figure~\ref{fig:saroo} shows face images of Saroo before he was lost and after he reunited with his biological mother. 
To understand the capability of face recognition technology to trace lost children, it is essential to systematically evaluate the longitudinal performance of face recognition technology on child face datasets.

While face recognition systems have improved the recognition performance under factors such as facial pose, illumination, and expression~\cite{frvt},~\cite{iarpa},~\cite{lfw},~\cite{blufr}, issues of aging and longitudinal studies~\footnote{Indeed, longitudinal studies for other biometric modalities are also limited, See Yoon and Jain~\cite{Yoon} for fingerprint study and Grother~\etal~\cite{Grother} for iris.} have not received adequate attention. Limited studies related to aging have indeed shown that (i) accuracy of face recognition degrades with an increase in time lapse between a subject's gallery and probe image acquisitions~\cite{klare-jain},~\cite{otto-jain},~\cite{ramanathan}, and (ii) face recognition accuracies for older subjects are higher than younger ones~\cite{ramanathan},~\cite{grother-frvt}.
 To the best of our knowledge, the largest longitudinal face datasets, consisting primarily of face images of adults\footnote{All subjects are above 18 years of age.}, are PCSO, LEO, and MSP\footnote{PCSO, LEO, and MSP are all operational face datasets and are not available in the public domain.} which were utilized in ~\cite{Lacey} and~\cite{Deb}. Deb~\etal~\cite{Deb} report that genuine scores of 99.0\% of the population remain above the threshold at a FAR of 0.01\% for an elapsed time of 10.5 years for a state-of-the-art COTS face matcher on both the PCSO and MSP datasets. However, these datasets are comprised of subjects above the age of $18$ and are not suitable for our study which focuses on tracing missing children. 

Prior studies on longitudinal face recognition performance is limited due to (i) lack of publicly available longitudinal face dataset of children, and (ii) low confidence in the accuracy of face recognition of children obtained by COTS matchers, which are primarily trained on adult face datasets. Best-Rowden~\etal studied face recognition performance of newborns, infants, and toddlers (ages $0$ to $4$ years) on $314$ subjects acquired over a maximum time lapse of only one year \cite{Lacey1}. Their results show that state-of-the-art face recognition technology has a very low True Accept Rate (TAR) of 47.93\% at 0.1\% False Accept Rate (FAR) for this age group of [0, 4] years. Based on their results, Best-Rowden~\etal suggested that longitudinal study of face recognition performance for faces enrolled at least 3 years of age or older may be feasible. Ricanek~\etal reviewed multiple face recognition algorithms on longitudinal face images from the In-the-Wild Child Celebrity (ITWCC)\footnote{ITWCC dataset is not in the public domain.} dataset, where the average age of subjects at enrollment is 10.2 years~\cite{ricanek}. A verification accuracy of 24\% at 0.1\% FAR was achieved, whereas closed-set identification performance was only 25\%.

To the best of our knowledge, the only two publicly available face image datasets that include children in the age group of $2$ to $18$ years are FG-NET~\cite{fgnet} and FaceTracer~\cite{facetracer}. FaceTracer has only one face image per child and FG-NET has only 400 images of subjects below the age of 15 years. The Cross-Age Celebrity Dataset (CACD)~\cite{cacd} was collected to evaluate face recognition performance under aging, but subjects younger than 10 years old are not included in this dataset, and only 199 subjects are present below the age of 18. Tables~\ref{tab:related} and~\ref{tab:allDatasets} concisely enumerate related works and longitudinal datasets~\footnote{Longitudinal data are repeated measurements on a collection of individual's sampled from a population over time. In contrast, cross-sectional data contains a single measurement made on each individual~\cite{msm}.}, respectively.

While no publicly available longitudinal datasets of children in the age range $[2, 18]$ years exists, we were able to obtain such a dataset, called Children Longitudinal Face\footnote{Due to privacy issues, CLF dataset cannot be released in the public domain. However, for repeatability studies, interested readers can obtain similarity scores computed using FaceNet from the authors.} (CLF) consisting of 3,682 face images of 919 subjects with an average of 4 images per subject collected over an average time span of 4.2 years. To the best of our knowledge, CLF is the largest longitudinal dataset in the aforementioned age group. 

Concisely, contributions of this paper are as follows:
\begin{enumerate}
\itemsep-1pt 
\item Evaluate the longitudinal performance of two state-of-the-art face recognition systems, COTS-A\footnote{Uses a convolutional neural network for face recognition} and FaceNet\footnote{The open-source face matcher, FaceNet, is available at~\url{https://github.com/davidsandberg/facenet}.}~\cite{facenet},~\cite{facenet-github} and a simple sum fusion of scores obtained from these two face matchers (referred to as Fused), on face images of children. To the best of our knowledge, no such longitudinal study exists for children in the age range of $2$ to $18$ years.
\item Formal statistical analysis of rates of change in face comparison scores obtained from COTS-A, FaceNet, and Fused face matchers due to covariates such as elapsed time between enrollment and probe images, and gender of the subjects.
\item Verification accuracy of 90.18\% at 0.1\% FAR is achieved by Fused after 1 year of time lapse between enrollment and probe image, which degrades to 73.33\% after 3 years of time lapse. Furthermore, Fused has a Rank-1 identification performance of 77.86\% at 1.0\% FAR after 1 year of elapsed time. We estimate that 80\% of the population in the CLF dataset can be successfully recognized at 0.1\% FAR by Fused over a gap of 2.5 years.
\end{enumerate}

The paper is organized as follows. Section~\ref{sec:dataset} details the longitudinal dataset used in this study. Section~\ref{sec:results} explains the experiments conducted in this study and outlines findings based on the experimental results. Section~\ref{sec:conclusion} concludes our paper and summarizes the results.

\section{Children Longitudinal Face (CLF) dataset}
\label{sec:dataset}

\begin{figure}[!t]
\centering
\begin{minipage}{0.24\linewidth}
\centering
\includegraphics[height=1in,keepaspectratio]{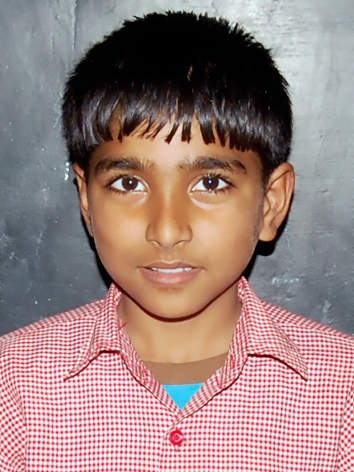}
\vfil
\footnotesize 9
\end{minipage}
\begin{minipage}{0.24\linewidth}
\centering
\includegraphics[height=1in,keepaspectratio]{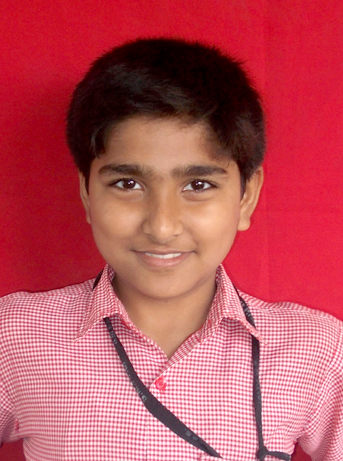}
\vfil
\footnotesize 11
\end{minipage}
\begin{minipage}{0.24\linewidth}
\centering
\includegraphics[height=1in,keepaspectratio]{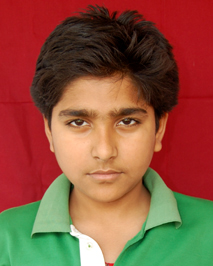}
\vfil
\footnotesize 12
\end{minipage}
\begin{minipage}{0.24\linewidth}
\centering
\includegraphics[height=1in,keepaspectratio]{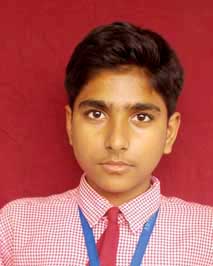}
\vfil
\footnotesize 14
\end{minipage}

\begin{minipage}{0.24\linewidth}
\centering
\includegraphics[height=1in,keepaspectratio]{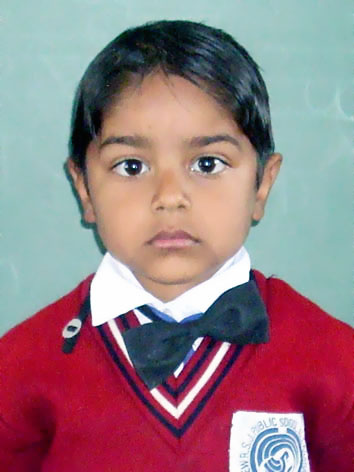}
\vfil
\footnotesize 4
\end{minipage}
\begin{minipage}{0.24\linewidth}
\centering
\includegraphics[height=1in,keepaspectratio]{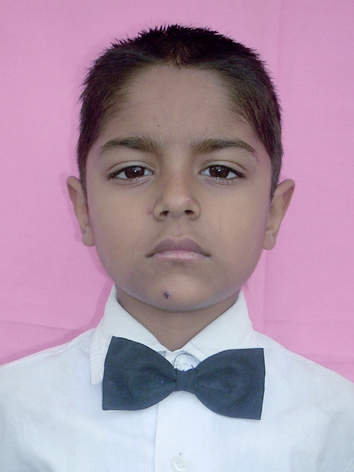}
\vfil
\footnotesize 5
\end{minipage}
\begin{minipage}{0.24\linewidth}
\centering
\includegraphics[height=1in,keepaspectratio]{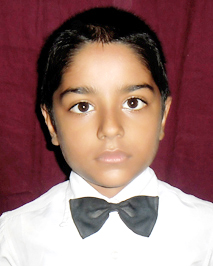}
\vfil
\footnotesize 7
\end{minipage}
\begin{minipage}{0.24\linewidth}
\centering
\includegraphics[height=1in,keepaspectratio]{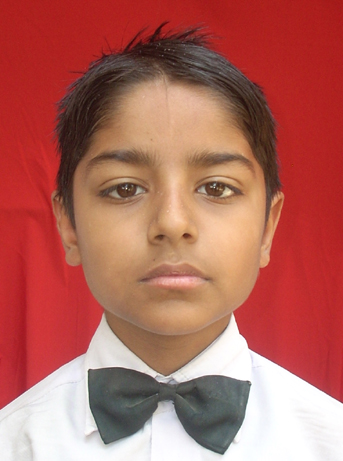}
\vfil
\footnotesize 8
\end{minipage}

\begin{minipage}{0.24\linewidth}
\centering
\includegraphics[height=1in,keepaspectratio]{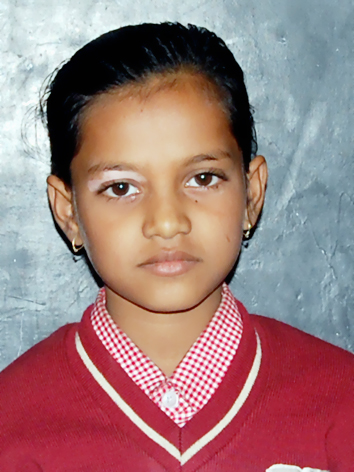}
\vfil
\footnotesize 10
\end{minipage}
\begin{minipage}{0.24\linewidth}
\centering
\includegraphics[height=1in,keepaspectratio]{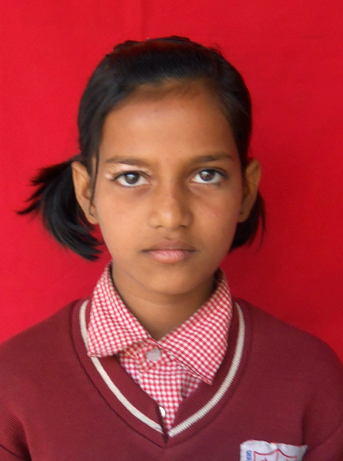}
\vfil
\footnotesize 12
\end{minipage}
\begin{minipage}{0.24\linewidth}
\centering
\includegraphics[height=1in,keepaspectratio]{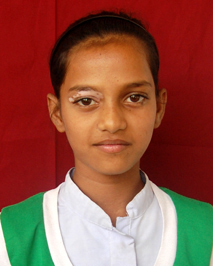}
\vfil
\footnotesize 13
\end{minipage}
\begin{minipage}{0.24\linewidth}
\centering
\includegraphics[height=1in,keepaspectratio]{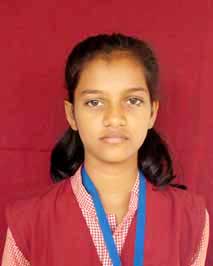}
\vfil
\footnotesize 15
\end{minipage}

\begin{minipage}{0.24\linewidth}
\centering
\includegraphics[height=1in,keepaspectratio]{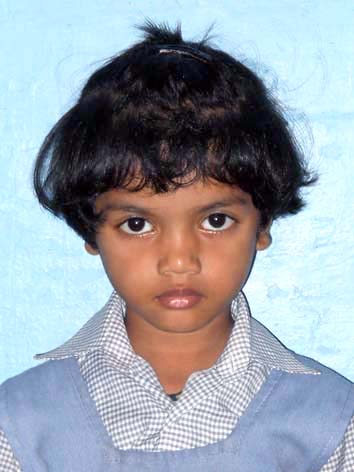}
\vfil
\footnotesize 5
\end{minipage}
\begin{minipage}{0.24\linewidth}
\centering
\includegraphics[height=1in,keepaspectratio]{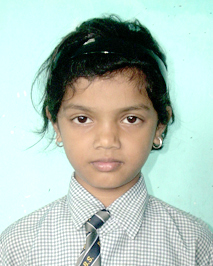}
\vfil
\footnotesize 8
\end{minipage}
\begin{minipage}{0.24\linewidth}
\centering
\includegraphics[height=1in,keepaspectratio]{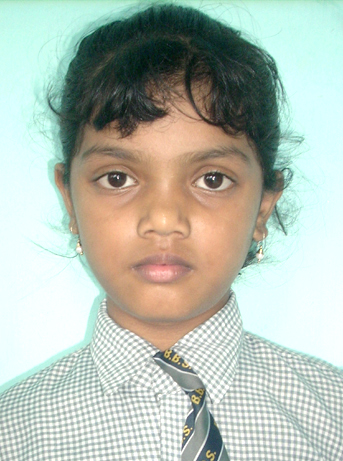}
\vfil
\footnotesize 9
\end{minipage}
\begin{minipage}{0.24\linewidth}
\centering
\includegraphics[height=1in,keepaspectratio]{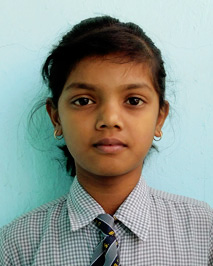}
\vfil
\footnotesize 10
\end{minipage}

\caption{Examples of longitudinal face data of four subjects (one row per subject), where images were acquired annually, in the CLF dataset. Age at image acquisition (in years) is given below each image.} 
\label{fig:samples}
\end{figure}

\begin{figure}[!t]
\centering
\begin{tabular}[p]{c}%
\begin{minipage}{0.19\linewidth}
\centering
\includegraphics[height=0.8in,keepaspectratio]{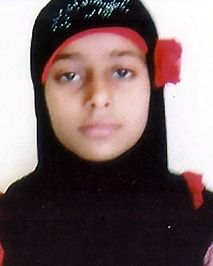}
\vfil
\footnotesize 10
\end{minipage}
\begin{minipage}{0.19\linewidth}
\centering
\includegraphics[height=0.8in,keepaspectratio]{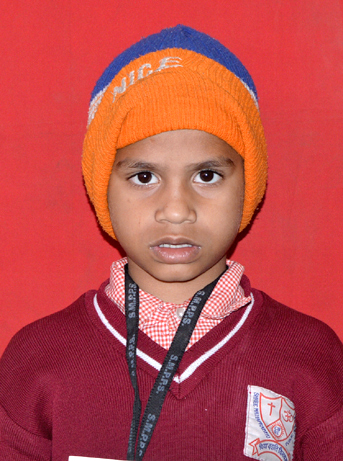}
\vfil
\footnotesize 9
\end{minipage}
\begin{minipage}{0.19\linewidth}
\centering
\includegraphics[height=0.8in,keepaspectratio]{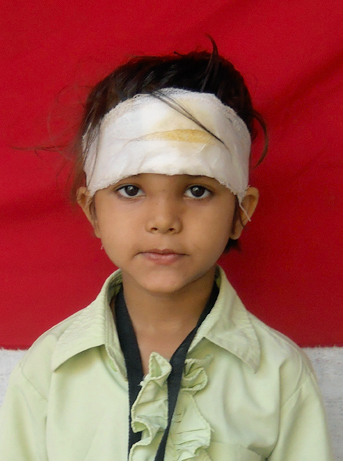}
\vfil
\footnotesize 5
\end{minipage}
\begin{minipage}{0.19\linewidth}
\centering
\includegraphics[height=0.8in,keepaspectratio]{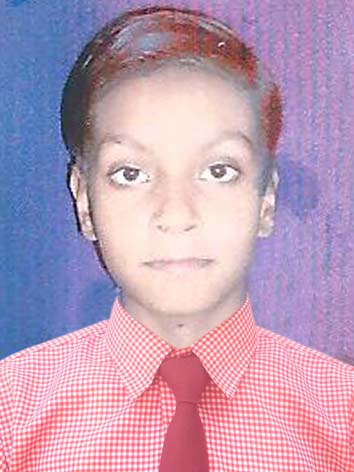}
\vfil
\footnotesize 9
\end{minipage}\\
\\
\begin{minipage}{0.23\linewidth}
\centering
\includegraphics[height=0.8in,keepaspectratio]{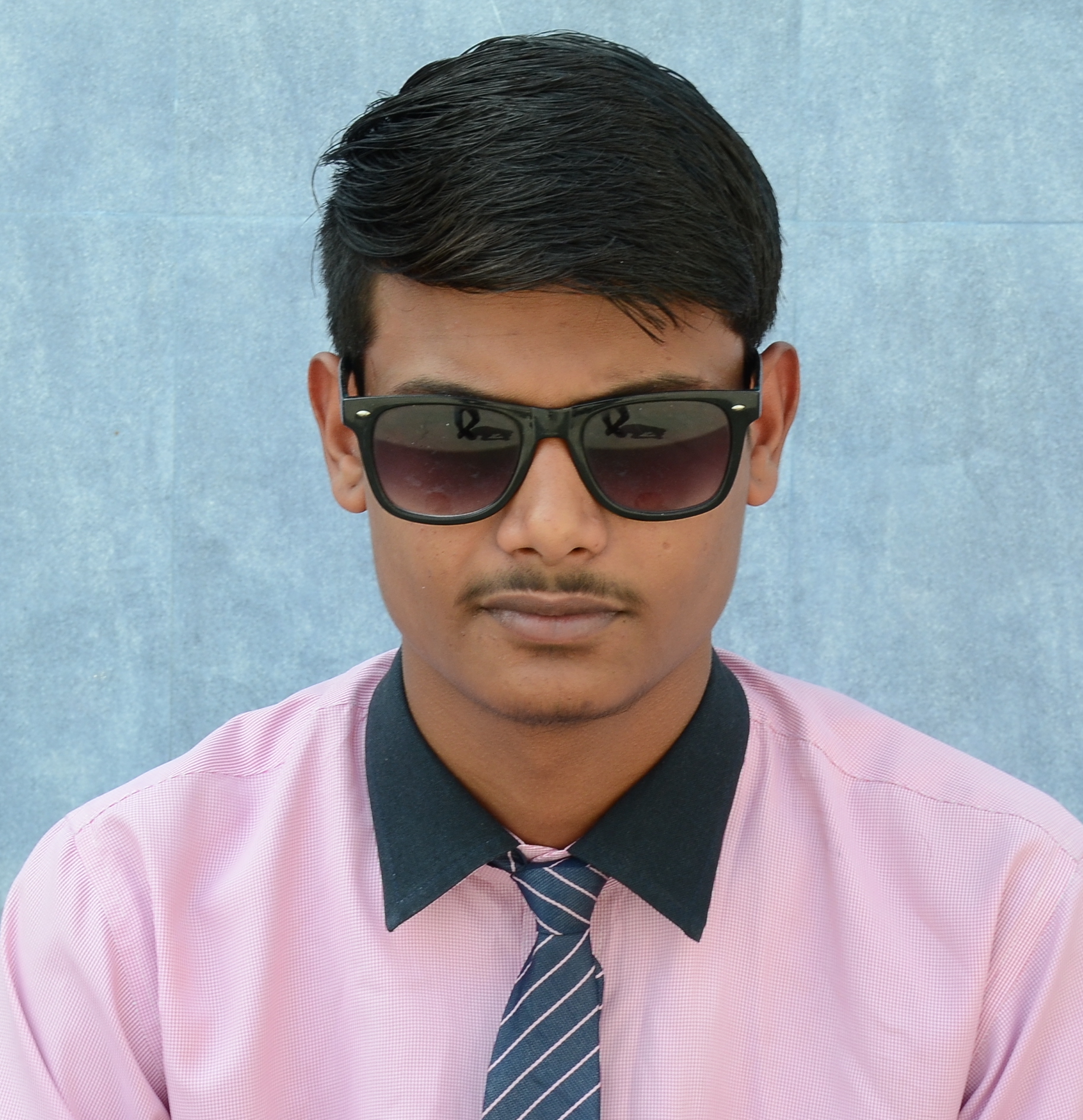}
\vfil
\footnotesize 15
\end{minipage}
\begin{minipage}{0.19\linewidth}
\centering
\includegraphics[height=0.8in,keepaspectratio]{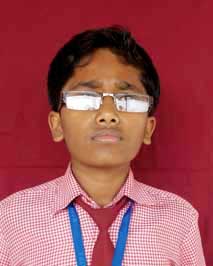}
\vfil
\footnotesize 13
\end{minipage}
\begin{minipage}{0.19\linewidth}
\centering
\includegraphics[height=0.8in,keepaspectratio]{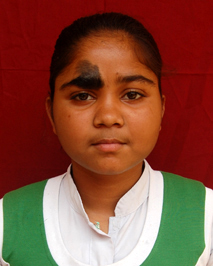}
\vfil
\footnotesize 14
\end{minipage}
\begin{minipage}{0.18\linewidth}
\centering
\includegraphics[height=0.8in,keepaspectratio]{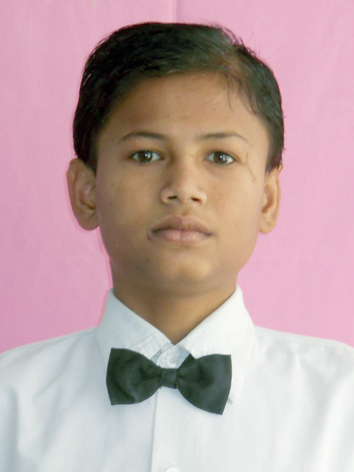}
\vfil
\footnotesize 11
\end{minipage}\\
\end{tabular}
\caption{CLF dataset examples with pose, illumination and expression variations, occlusions due to head covering, cap, bandage, beard, and sunglasses, and moles and scars. Age at image acquisition (in years) is given below each image.} 
\label{fig:bad}
\end{figure}

\begin{figure}[!t]
\subfloat[]{\resizebox{0.5\linewidth}{!}{\includegraphics{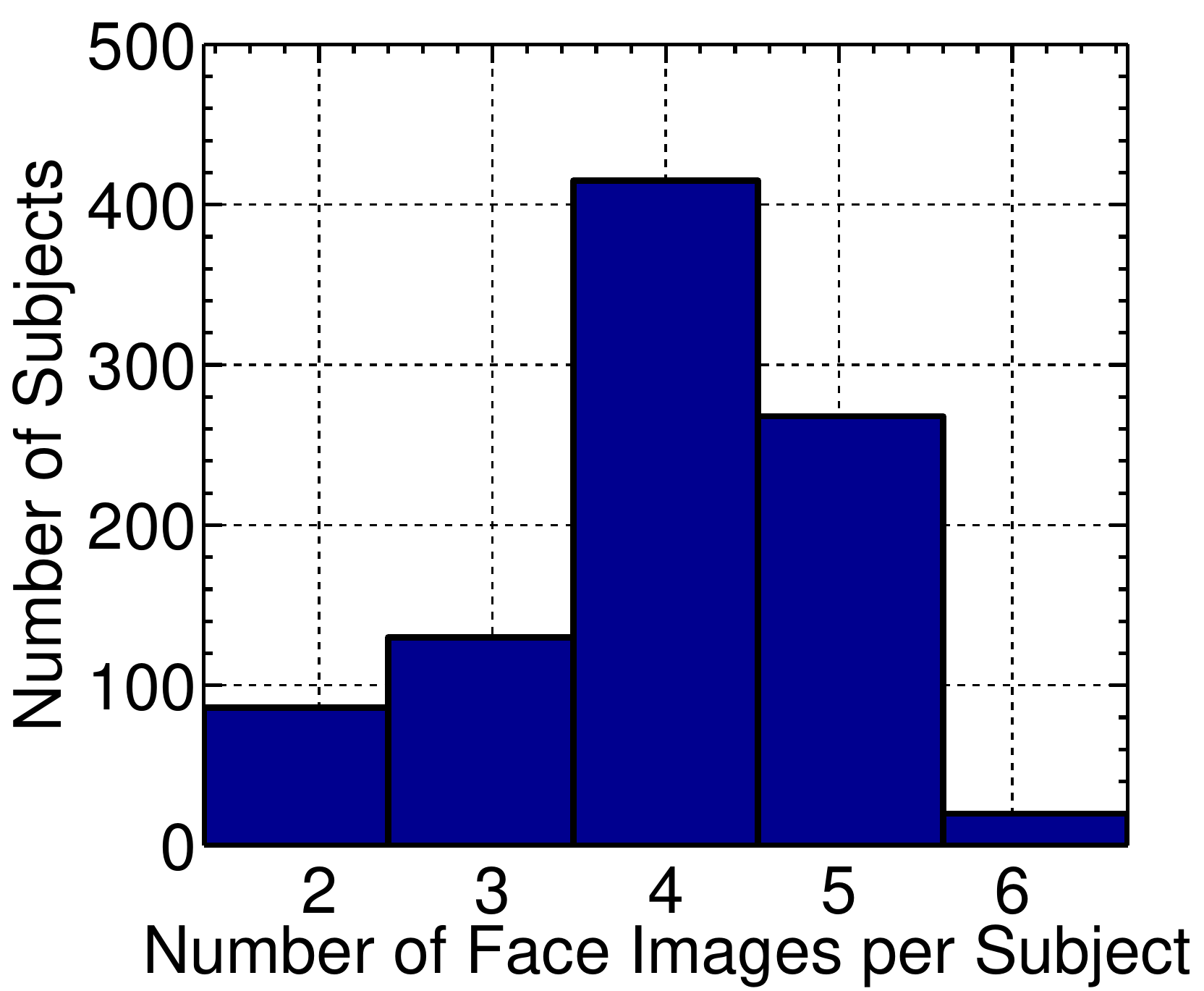}}}
\subfloat[]{\resizebox{0.5\linewidth}{!}{\includegraphics{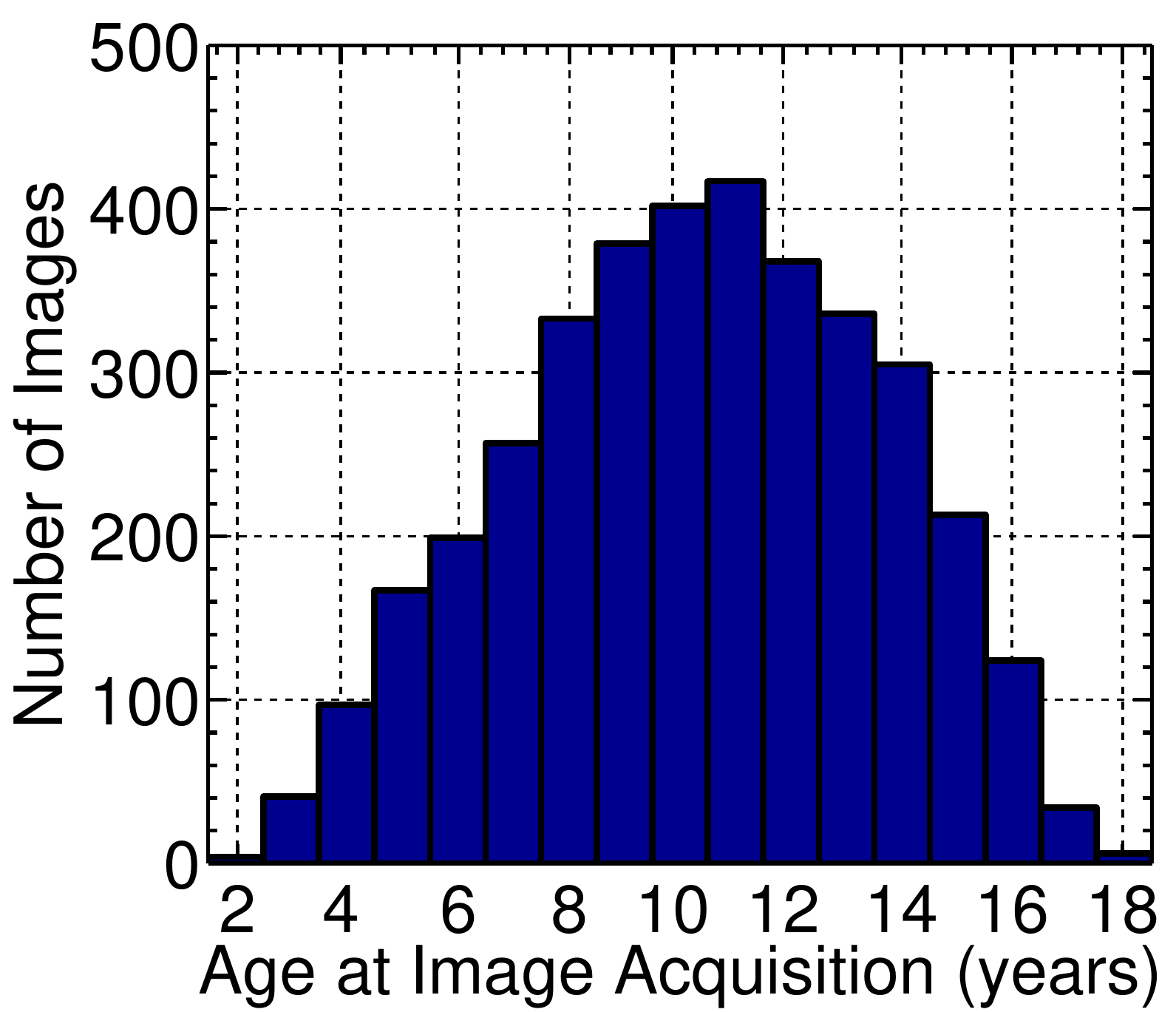}}}
\vfil
\subfloat[]{\resizebox{0.5\linewidth}{!}{\includegraphics{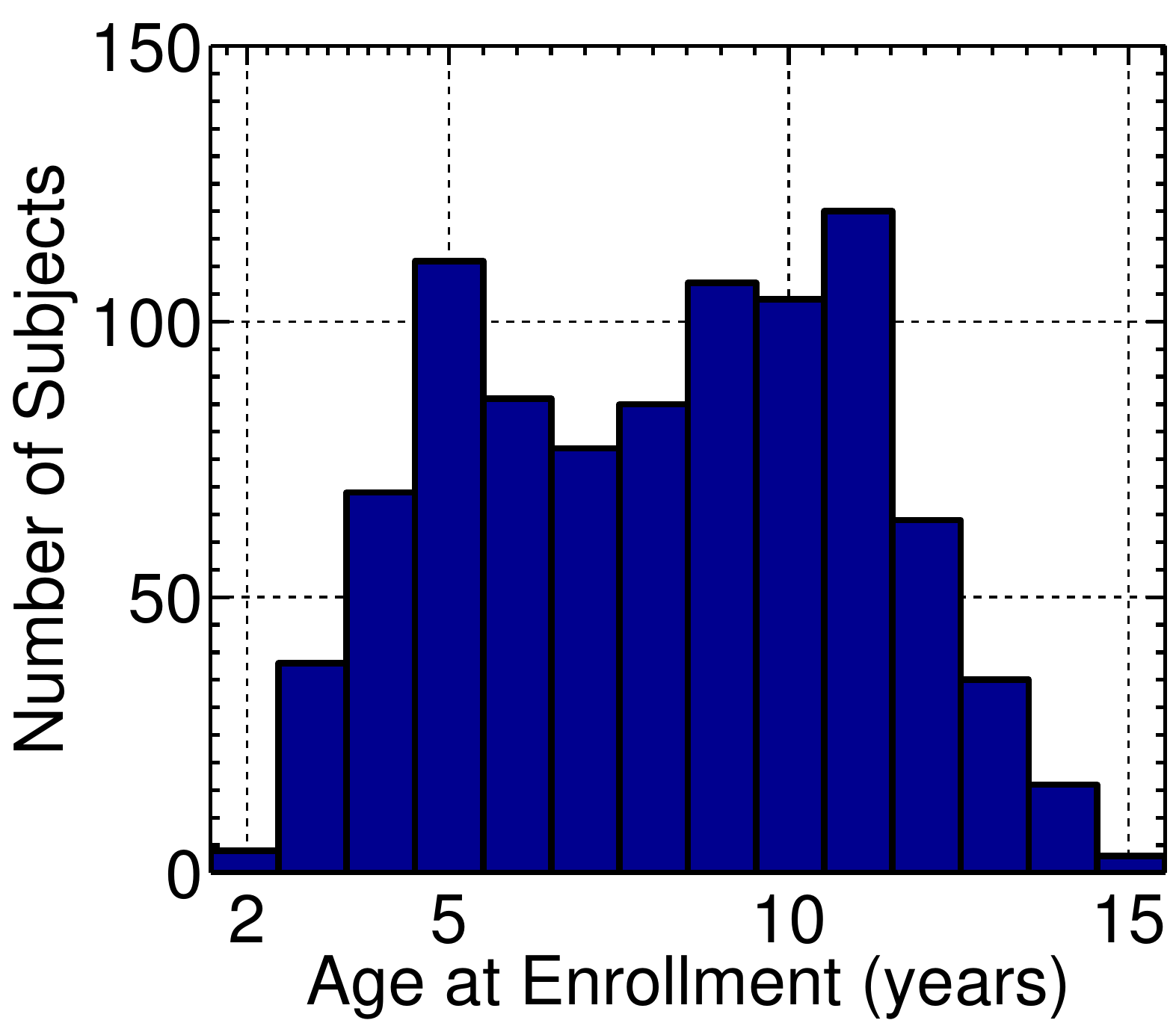}}}
\subfloat[]{\resizebox{0.5\linewidth}{!}{\includegraphics{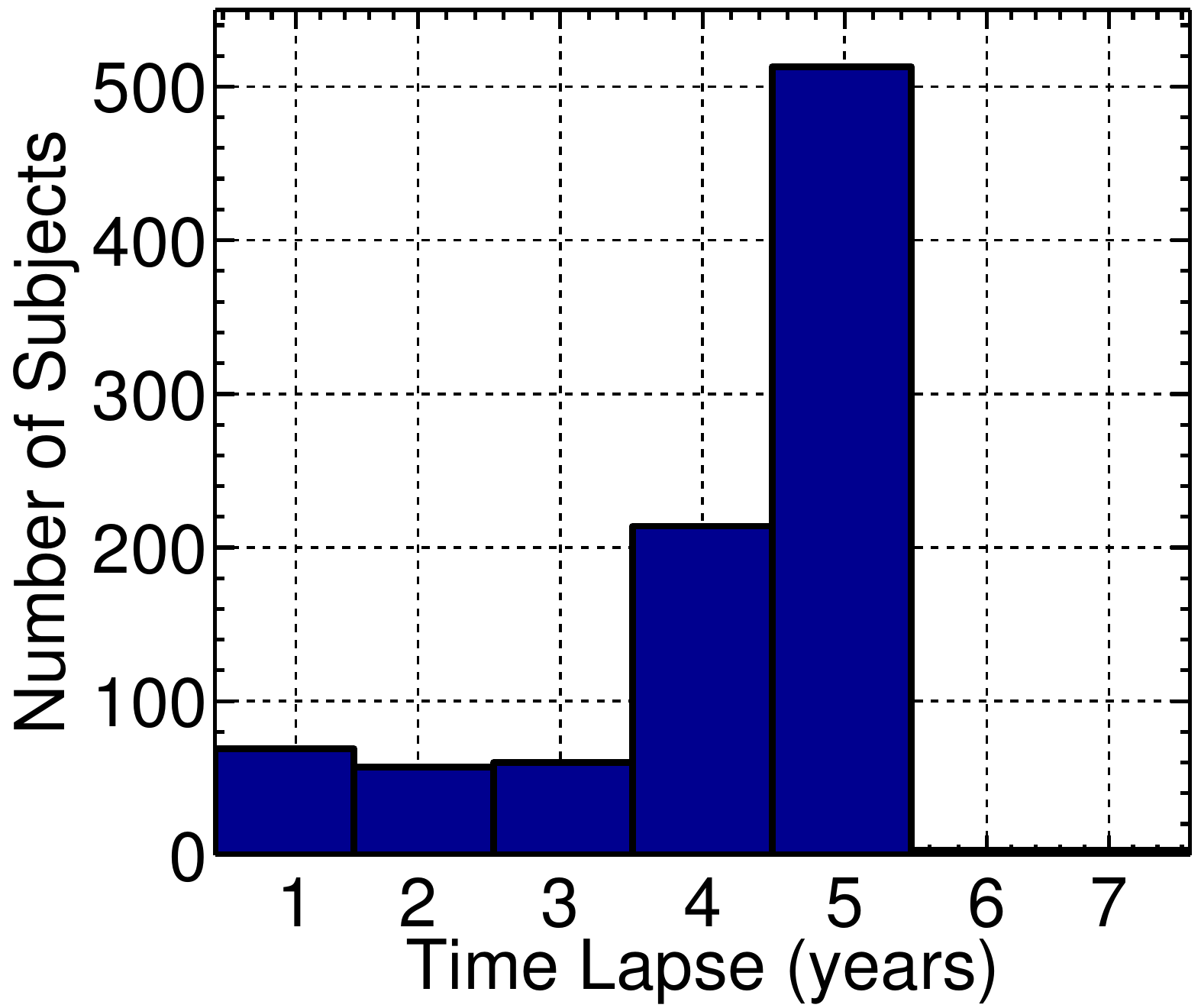}}}
 \caption{CLF statistics: a) number of face images per subject; b) age at image acquisitions in years; c) enrollment age of a subject in years; d) time span between enrollment and latest image acquisition for each subject in years.}
\label{fig:statistics}
\end{figure}

The Children Longitudinal Face (CLF) dataset contains $3,682$ face images of $919$ children, in the age range of $2$ to $18$ years. Each subject has an average of $4$ images acquired over an average time lapse of 4 years (minimum time lapse of $2$ years; maximum time lapse of $7$ years). Demographic makeup of CLF dataset is comprised of $604$ ($66\%$) boys and $315$ ($34\%$) girls. Dataset statistics are shown in Figure~\ref{fig:statistics}. The face images were captured with a resolution of $354\times472$ pixels (Figure~\ref{fig:samples}). Figure~\ref{fig:bad} shows example image acquisitions with challenging variations in i) pose, illumination and expression, ii) obstructions such as scarves, cap, bandage, beard, and spectacles, and iii) birth marks such as moles, cuts, distinct eye color, and scars. Due to zoom variations, some faces occupy about only $70\%$ of the image while some faces cover about $50\%$ of the total image area. 

The following criteria were used to postprocess the dataset: 
\begin{itemize}
\itemsep-1pt 
\item Each subject has only one image acquisition session
\item De-duplication of identities
\item Date of birth for each subject was recorded at each session. In case of a missing date of birth for a session, we used the date of birth recorded at the time of enrollment to estimate the subject's age at the session.
\end{itemize}

\section{Experiments}
\label{sec:results}
\begin{figure}[!t]
\centering
\includegraphics[width=0.7\linewidth,keepaspectratio]{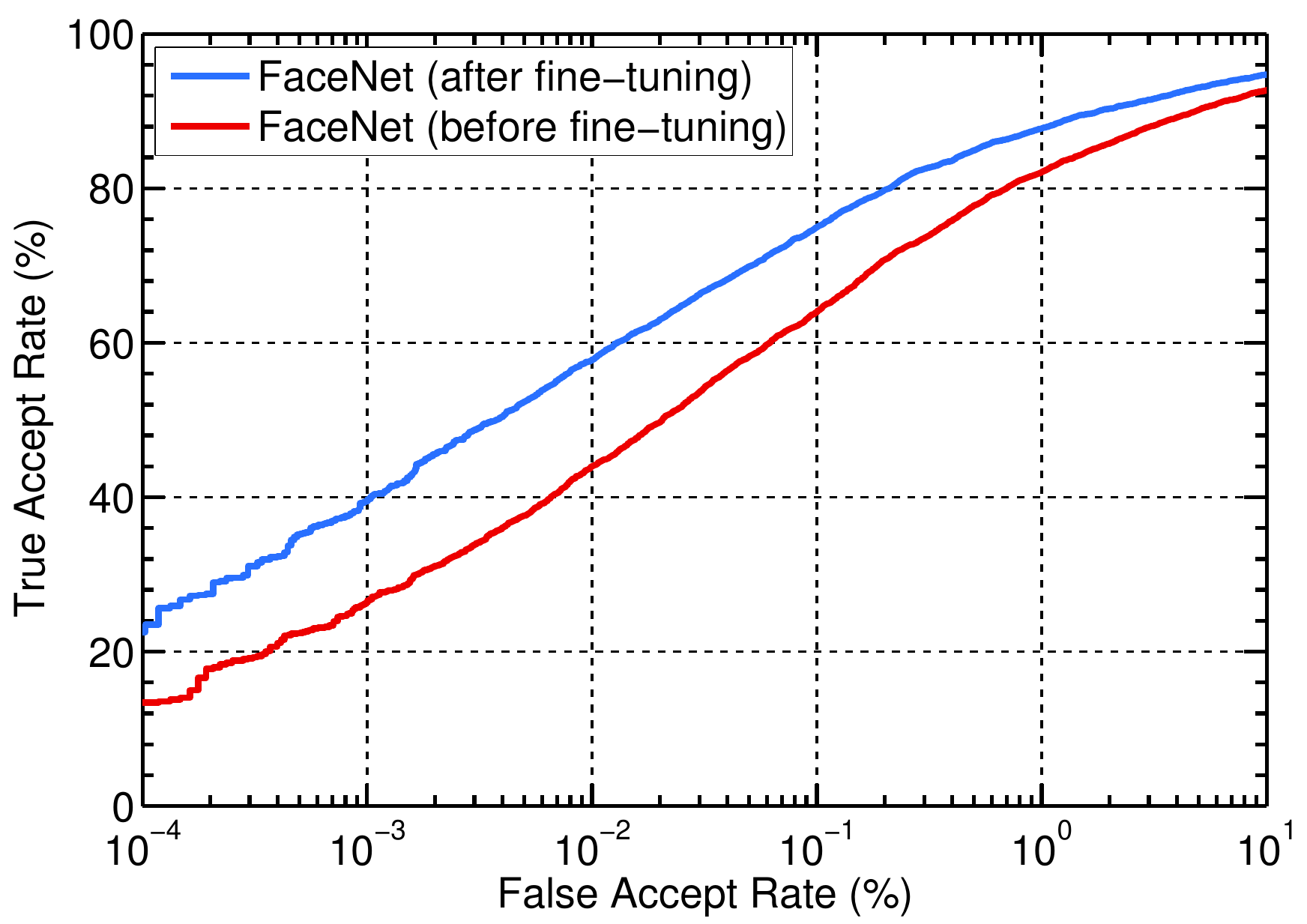}
 \caption{Face recognition performance of FaceNet before and fine-tuning on a child face training training dataset shown in red and blue, respectively. After fine-tuning, performance is significantly improved.}
\label{fig:compareOldNewFN}
\end{figure}

Performance of two state-of-the-art face recognition systems, COTS-A and FaceNet, are evaluated on Children Longitudinal Face (CLF) dataset. In addition, performance of sum score fusion of the above two face matchers, Fused, is also reported. FaceNet is originally trained on a publicly available dataset, MS-Celeb\footnote{MS-Celeb dataset can be downloaded from~\url{https://www.microsoft.com/en-us/research/project/ms-celeb-1m-challenge-recognizing-one-million-celebrities-real-world}}~\cite{ms_celeb}, comprising of 10 million face images of 100K celebrities. Face images in the dataset are acquired by leveraging public search engines to provide approximately 100 images per celebrity. Face recognition accuracy of the FaceNet matcher on CLF dataset is quite low (43.87\% TAR at 0.01\% FAR) because it was trained on adult faces. To boost face recognition performance, we fine-tuned FaceNet on an independent set of 3,294 face images of 1,119 children in the age group $3$ to $18$ years (different dataset than the CLF dataset), denoted as Child Face Training (CFT) dataset. For both the FaceNet models (before and after fine-tuning), feature vectors (128-dimensional) for all face images in the CLF dataset are extracted and face comparison scores are obtained by the cosine-similarity metric. Genuine scores (total of 5,946 scores) are computed as all pairwise comparisons between face images of the same subject and impostor scores are comprised of all possible impostor comparisons (total of 3.38 million scores) in the CLF dataset. Figure~\ref{fig:compareOldNewFN} shows that the performance of FaceNet is significantly improved after fine-tuning it on the CFT dataset. FaceNet achieves TARs of $43.87\%$ and $57.74\%$, both at $0.01\%$ FAR, with the original model and the fine-tuned model, respectively. Therefore, only the fine-tuned FaceNet face matcher will be subsequently used in our study.

\subsection{Verification and Identification Scenarios}\label{sec:verification_identification}
For evaluating longitudinal performance of COTS-A, FaceNet, and Fused, face images for each subject at enrollment (first image acquisition) are compared to subsequent face image acquisitions of the same subject (a total of 2,763 genuine comparison scores). Longitudinal performance is evaluated after 1, 3, and 5 years of elapsed time since enrollment ($\Delta T$). In the verification scenario, the impostor distribution includes all possible impostor comparisons, totaling 3.38 million scores. Table~\ref{tab:veriOsi} provides verification accuracies at both 0.01\% and 0.1\% FAR values for an elapsed time of 3 and 5 years. We find that there is a decreasing trend in face recognition accuracy over time, which is consistent with findings in prior studies~\cite{klare-jain},~\cite{otto-jain},~\cite{ramanathan}. Fused face matcher has the best verification performance over time (90.18\% TAR at 0.1\% FAR with 1 year time lapse), compared to COTS-A (81.94\% TAR at 0.1\% FAR with 1 year time lapse) and FaceNet (83.77\% TAR at 0.1\% FAR with 1 year time lapse). The improved performance of Fused suggests that COTS-A and FaceNet matchers are complementary in nature\footnote{The training datasets for COTS-A and FaceNet are likely different which may account for an improved face recognition performance upon fusing their scores.}. 

\begin{table*}[t]
\footnotesize
\caption{Longitudinal performance of COTS-A, FaceNet and Fused for (a) verification and (b) open-set identification scenarios.}
\centering
\subfloat[TAR (\%) @ FAR]{\begin{tabularx}{0.51\linewidth}{|Y| Y| Y| Y| Y|}
\noalign{\hrule height 1.5pt}
Matcher & \multicolumn{2}{|c|}{$\Delta T = 1$ year} & \multicolumn{2}{|c|}{$\Delta T = 3$ years}\\
\noalign{\hrule height 1.5pt}
& 0.01\% FAR &  0.1\% FAR & 0.01\% FAR & 0.1\% FAR\\
\noalign{\hrule height 1.2pt}
 COTS-A & 64.51 & 81.94 & 38.18 & 49.33\\
   \hline
  FaceNet  & 67.94 & 83.77 & 34.75 & 59.80\\
  \hline
  Fused  & 80.56 & 90.18 & 53.33 & 73.33\\
\noalign{\hrule height 1.5pt}
\end{tabularx}}\hfil
\subfloat[DIR (\%) at 1\% FAR,]{\begin{tabularx}{0.47\linewidth}{|Y| Y| Y| Y| Y|}
\noalign{\hrule height 1.5pt}
Matcher & \multicolumn{2}{|c|}{$\Delta T = 1$ year} & \multicolumn{2}{|c|}{$\Delta T = 7$ years}\\
\noalign{\hrule height 1.5pt}
& Rank-1 &  Rank-3 & Rank-1 & Rank-3\\
\noalign{\hrule height 1.2pt}
 COTS-A & 68.49 & 70.65 & 66.89 & 68.59\\
   \hline
  FaceNet  & 55.31 & 55.89 & 51.73 & 52.42\\
  \hline
  Fused  & 77.86 & 79.01 & 75.39 & 76.42\\
\noalign{\hrule height 1.5pt}
\end{tabularx}
\label{iden}}
\label{tab:veriOsi}
\end{table*}

In the identification scenario, we keep all 919 enrollment images for all the subjects in the CLF dataset in the gallery and the non-enrollment image acquisitions in the probe set. Additionally, we included 756 subjects with one face image per subject (not in the CLF dataset) in our probe set for open-set identification, totaling 3,520 probe images. Rank-1 and Rank-3 identification accuracies are computed at 1 and 7 years of elapsed time (Table~\ref{iden}). Similar to the verification scenario, identification performance decreases with an increase in time lapse, however, the rate of degradation in identification accuracy over time is very low. This suggests that identification of missing children is feasible over a time lapse of 7 years between a child's enrollment image in the gallery and probe image. Detection and Identification Rate (DIR) remains stable at ranks beyond 3 for all face recognition systems, which seems to suggest that if a subject is not found within the first three ranks, it is unlikely that the subject will be identified at a higher rank.

\subsection{Multilevel Statistical Models}
A longitudinal analysis of genuine scores for child face images is necessary to understand the variation in genuine scores over time and the impact of additional covariates, such as gender. Time lapse between a probe and enrollment image and number of image acquisitions per subject in the CLF dataset varies from subject to subject and therefore, the dataset is time-unstructured and unbalanced. Multilevel statistical models are recommended for analyzing such datasets where variations occur at different levels in the data hierarchy. Open-set identification relies on two tasks: verification and identification. A probe first claims to be present in the gallery, and a pre-determined threshold is used to accept or reject the claim using similarity scores (verification). If the probe is accepted, the ranked list of gallery images which match the probe with similarity scores above the threshold are returned as the candidate list (identification). Longitudinal analysis in this section is conducted in the verification scenario to first analyze the magnitudes of genuine similarity scores over time and determine the impact on the verification task of open-set identification.
\begin{table*}[t]
\footnotesize
\caption{Multilevel models with different covariates}
\centering
\resizebox{0.85\linewidth}{!}{
\begin{tabular}{c c c c}
\noalign{\hrule height 1.5pt}
Model & Level-1 Model & Level-2 Model & Covariates\\
   \hline
 Model $B_{T}$ & $Y_{ij}$ = $\pi_{0i}$ + $\pi_{1i}\Delta T_{i,j}$ + $\varepsilon_{i,j}$ &  $\pi_{0i}$ = $\gamma_{00}$ + $b_{0i}$, & Time lapse \\&&$\pi_{1i}$ = $\gamma_{10}$ + $b_{1i}$ \\
   \hline
    Model $C_{Gender}$ & $Y_{ij}$ = $\pi_{0i}$ + $\pi_{1i}\Delta T_{i,j}$ + $\varepsilon_{i,j}$ &  $\pi_{0i}$ = $\gamma_{00}$ + $\gamma_{01}Gender_{i}$  + $b_{0i}$ & Time lapse, and gender \\&&$\pi_{1i}$ = $\gamma_{10}$ + $\gamma_{11}Gender_{i}$ + $b_{1i}$\\
\noalign{\hrule height 1.5pt}
\end{tabular}
}
\label{models}
\end{table*}

Let $N_{i}$ represent the total number of face image acquisitions for a child $i$ in the CLF dataset. If $I_{i,j}$ is the $j^{th}$ face image of child $i$, then $I_{i} = \{I_{i,0},I_{i,1},\ldots I_{i,N_{i}-1}\}$ represents the set of all $N_{i}$ image acquisitions for the child $i$. The set $I_{i}$ is ordered with increasing age at image acquisition. In other words, if $AGE_{i,j}$ gives the age at $j^{th}$ image acquisition of child $i$, then $AGE_{i,j} < AGE_{i,k}$ for $j=0,1,\ldots,N_{i}-2$ and $k = j+1,\ldots,N_i-1$. Genuine scores are obtained by comparing a child's enrollment image (first acquisition) to every other image acquisition, totaling $N_{i} - 1$ genuine scores for each subject $i$ in the dataset. The time lapse between a subject's enrollment image, $I_{i,0}$, and a query image, $I_{i,j}$ where $0<j\leq N_i - 1$, is given by $\Delta T_{i,j} = AGE_{i,j} - AGE_{i,0}$. $Y_{i,j}$, where $0<j\leq N_i - 1$,  represents the genuine comparison score between $j^{th}$ face image acquisition and enrollment image for a child $i$.

\begin{figure*}[!t]
\resizebox{0.33\textwidth}{!}{\includegraphics{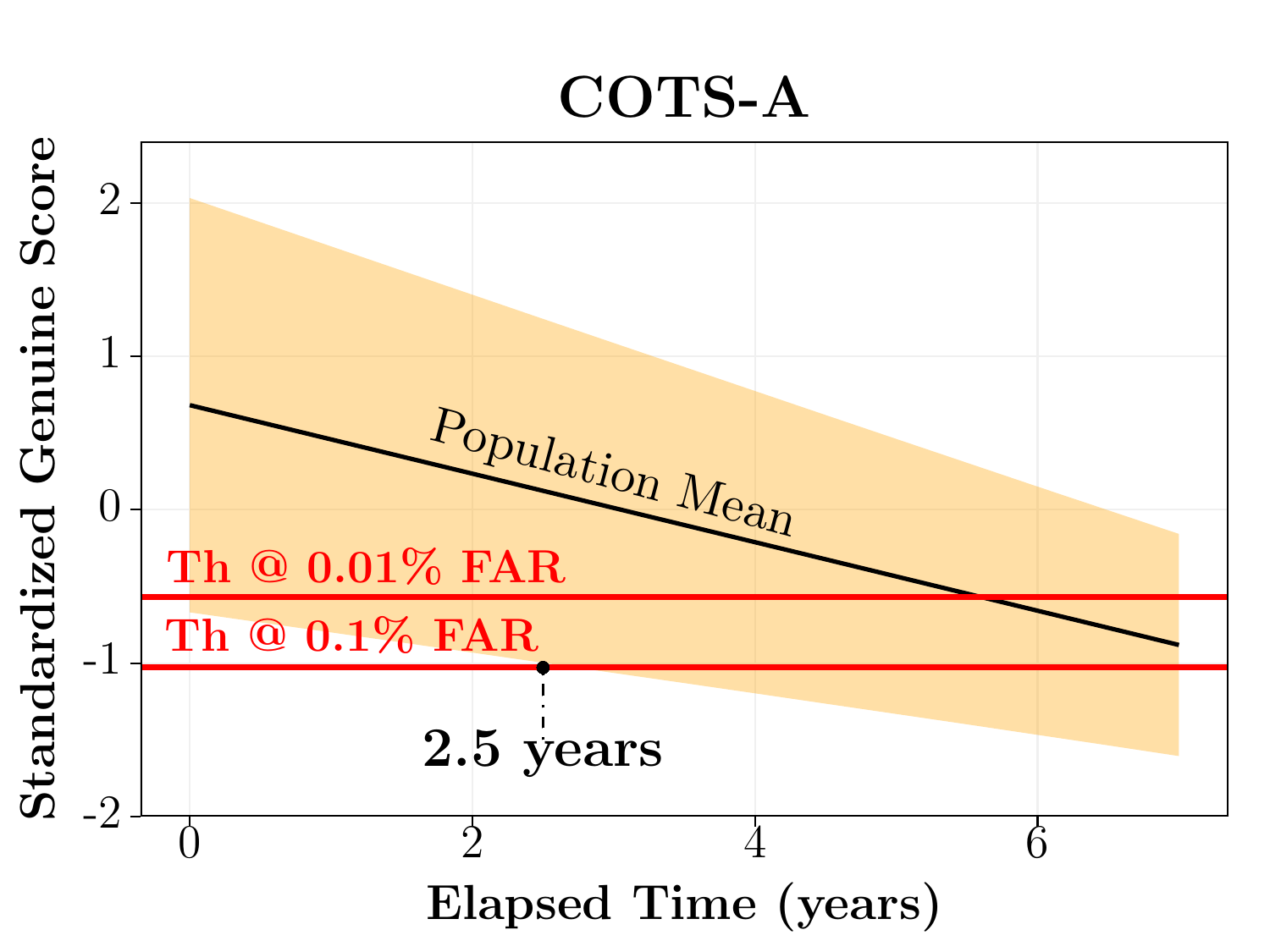}}
\resizebox{0.33\textwidth}{!}{\includegraphics{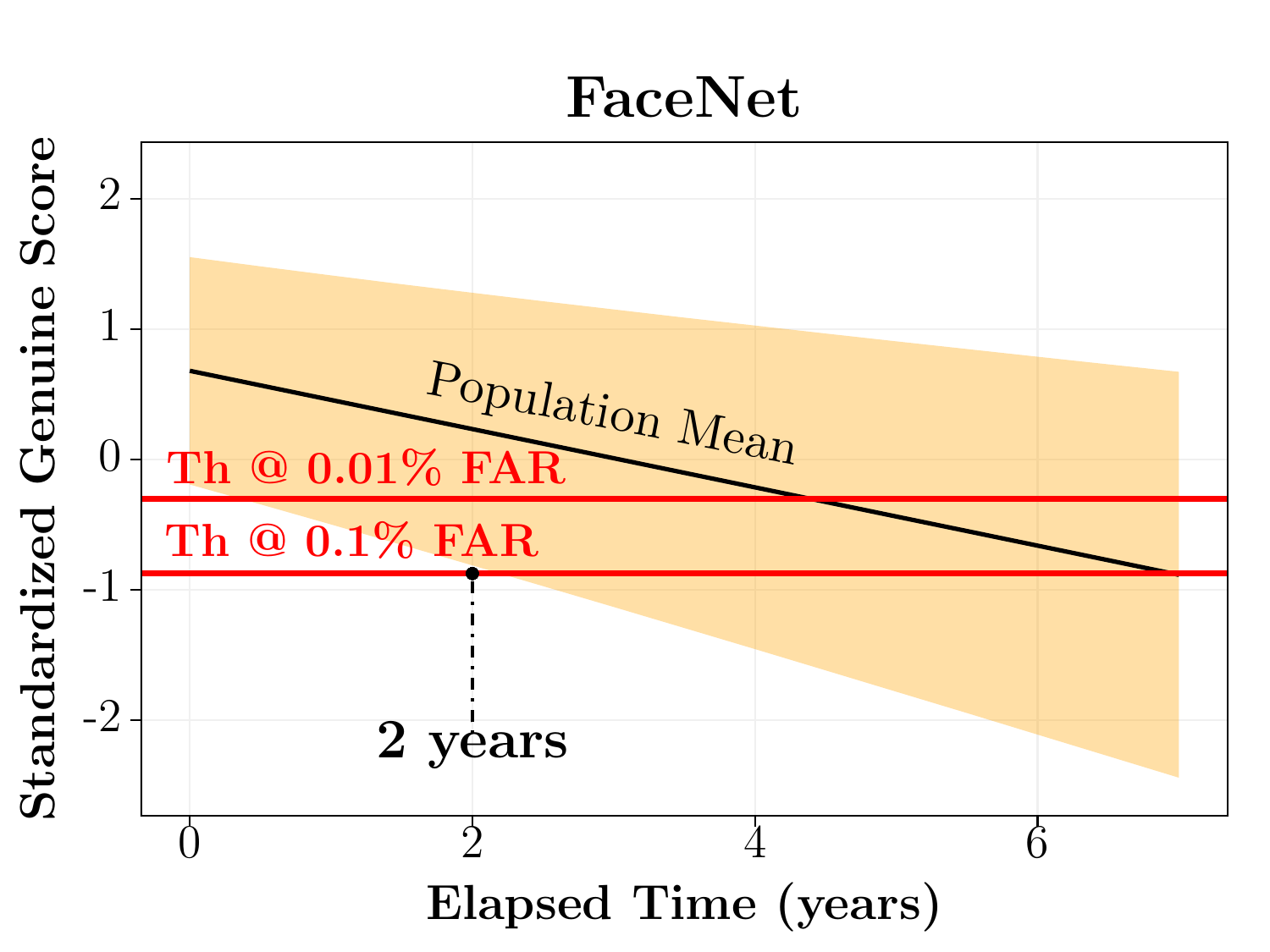}}
\resizebox{0.33\textwidth}{!}{\includegraphics{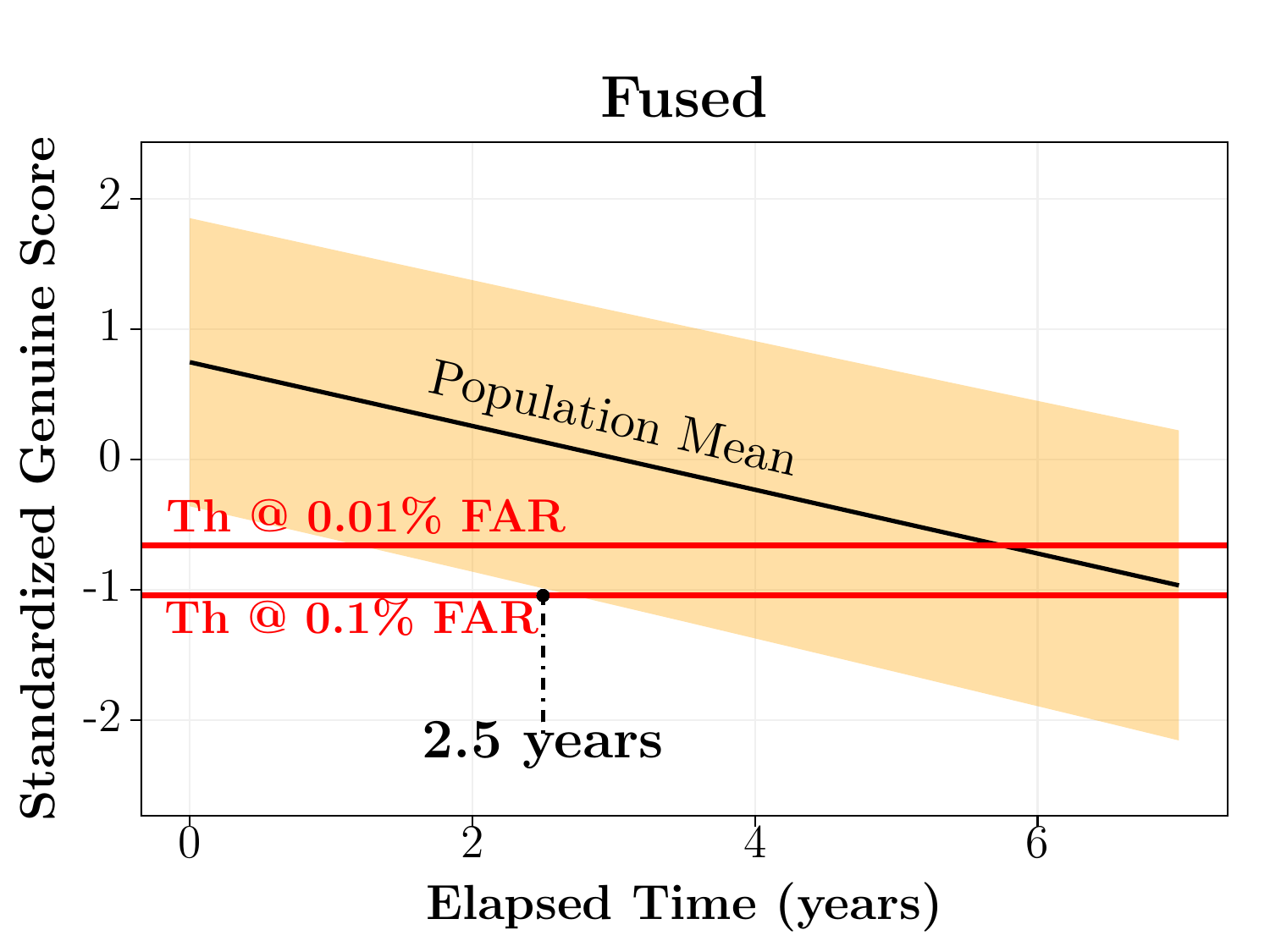}}%
 \caption{Results from fitting genuine scores obtained from COTS-A, FaceNet, and Fused face matchers to Model $B_T$ on the CLF dataset. The orange band estimates the regions containing longitudinal trends for 80\% of the CLF dataset population around the population-mean trend. Thresholds at 0.01\% and 0.1\% FAR for all three face matchers are also plotted in red lines.}
\label{timelapse}
\end{figure*}

Models used in this work are described using two hierarchical levels, similar to those described in~\cite{Deb},~\cite{Lacey}. The first level in the hierarchy, Level-1, models the changes in genuine scores, $Y_{i,j}$, for each subject over time (within-subject variation), whereas, Level-2 model accounts for variation in genuine scores across different subjects (between-subject variation). To quantify change in standard deviations of the genuine score distribution per year, genuine comparison scores are normalized such that $Y_{i,j} = \left(y_{i,j} - \mu\right)/\sigma$, where $y_{i,j}$ is the raw comparison score obtained from the face matchers, and $\mu$ and $\sigma$ are the mean and standard deviation of the genuine scores from all the subjects in the dataset. The trend in genuine scores over time is modeled as a linear function of various covariates, $X_{i,j}$,
\setlength{\abovedisplayskip}{0pt}%
\setlength{\belowdisplayskip}{0pt}%
\setlength{\abovedisplayshortskip}{0pt}%
\setlength{\belowdisplayshortskip}{0pt}%

\begin{align*}
Y_{i,j} = \pi_{0i} + \pi_{1i}X_{i,j} + \epsilon_{i,j}
\end{align*}
where $\pi_{0i}$ and $\pi_{1i}$ are subject $i$'s intercept and slope, respectively. This corresponds to our Level-1 model which models the within-subject changes in face comparison scores over time. Subject $i$'s face comparison scores can vary around his/her trend by $\epsilon_{i,j}$, the Level-1 residual variance. The slope and intercept parameters are a combination of \emph{fixed}, $\gamma_{00}, \gamma_{10}$, and \emph{random}, $b_{0i}, b_{1i}$, effects. Fixed effects are the overall means of the population intercepts and slopes, whereas, random effects are subject $i$'s deviation from the population means. Hence, $\pi_{0i}$ and $\pi_{1i}$. can be expanded to,
\begin{align*}
\pi_{0i} &= \gamma_{00}+b_{0i}\\
\pi_{1i} &= \gamma_{10}+b_{1i}
\end{align*}
corresponding to the Level-2 model. Therefore, our multilevel statistical model for the genuine score between subject $i$'s enrollment and $j^{th}$ image is simply,
\begin{align*}
Y_{i,j} = \left(\gamma_{00}+b_{0i}\right) + \left(\gamma_{10}+b_{1i}\right)X_{i,j} +  \epsilon_{i,j}.
\end{align*}
The following covariates for which we have the data are used in this study:
\begin{itemize}
\itemsep0em 
\item $\Delta T_{i,j}$: time lapse between a child $i$'s $j^{th}$ image acquisition and enrollment image
\item $Gender_{i}$: gender of child $i$ (0 for girl, 1 for boy)
\end{itemize}
Time lapse ($\Delta T_{i,j}$) affects our Level-1 model, whereas, gender ($Gender_{i}$) is time-invariant and affects between-subject variation (Level-2) model. Table~\ref{models} describes the models and covariates incorporated in this study.

Standardized genuine scores from COTS-A, FaceNet, and Fused are obtained, totaling 2,763 scores. To evaluate longitudinal accuracies, trends in genuine scores should be considered in context with an impostor distribution. For the CLF dataset, all possible impostor scores (3.38 million) are computed to calculate the thresholds at fixed FAR values. Longitudinal trends in genuine scores affecting the face recognition accuracies of the three face recognition systems are evaluated at thresholds corresponding to 0.01\% and 0.1\% FAR. 

Multilevel statistical models are based on the assumption that the residual errors are normally distributed. CLF dataset violates this parametric assumption of normality and therefore, non-parametric bootstrapping is performed to obtain confidence intervals for the parameter estimates~\cite{Yoon}. By sampling all the 919 subjects in the dataset with replacement, non-parametric bootstrapping is conducted with 1,000 bootstrap sets. The multilevel statistical models described in Table~\ref{models} are then fit, with the LME4 package in R using maximum likelihood estimation, to each bootstrap set and mean parameter estimates over all 1,000 bootstraps are computed. 

\label{sec:longitudinal}
\subsubsection{Time Lapse}
Model $B_T$ contains a covariate, $\Delta T_{i,j}$, which describes the time lapse between between a subject's enrollment image and probe image.
The population-mean trend, $\gamma_{00}, \gamma_{10}$, for Model $B_T$ estimates that COTS-A, FaceNet, and Fused genuine scores decrease by 0.2234, 0.2180, and 0.2444  standard deviations per year for CLF dataset, respectively. Therefore, genuine scores for COTS-A, FaceNet, and Fused decrease by one full standard deviation of their respective score distribution after 4.5, 4.6, and 4.1 years of time lapse.

Following the studies conducting in~\cite{Deb},~\cite{Lacey}, regions containing longitudinal trends for 80\%\footnote{Instead of analyzing at 95\% and 99\% confidence levels, 80\% is used because the face matchers did not achieve verification accuracies above 91\% at 0.1\% FAR, found in Section~\ref{sec:verification_identification}.} of the child population are plotted using estimated changes in slope and intercept parameters ($\sigma_0^2, \sigma_1^2, \sigma_{01}$). The regions are then used to determine the time lapse until genuine scores for 95\% and 99\% of the population begin to drop below thresholds at 0.01\% and 0.1\% FAR. Therefore, we estimate the elapsed time in years over which face recognition performance is stable before a decrease in genuine scores result in false accept errors. Figure~\ref{timelapse} suggests that genuine scores of 99\% of the population remain above the threshold at 0.01\% FAR for an elapsed time of 2.5, 2, and 2.5 years for COTS-A, FaceNet, and Fused face matchers, respectively, on the CLF dataset. We estimate that 80\% of the population in the CLF dataset can be successfully verified at 0.1\% FAR for up to 2.5 years, and Table~\ref{tab:veriOsi} found that the verification accuracy for Fused decreased from 90.18\% to 73.33\% for a time lapse of 3 years. 


\subsubsection{Gender}
We investigate whether variability in subject-specific longitudinal trends in genuine scores can be better explained by gender demographics. Population-mean trends for the gender model, $C_{Gender}$, for all the three face matchers have similar trends indicating that the effects of gender on the change in genuine scores for CLF dataset over time is matcher-independent. The average genuine scores were found to be not statistically different between boys and girls, however, the rates of change (slopes) is significantly steeper for boys than girls. Therefore, for all three face matchers, girls appear to be easier to recognize than boys with higher genuine scores overall. We suspect that the differences between boys and girls can be attributed to changes in facial hair for boys over time and possibly later maturity attained by boys~\cite{maturity}.

\section{Conclusions}
\label{sec:conclusion}
We investigated the performance of two state-of-the-art face recognition systems and their fusion for child face recognition in the age group $[2, 18]$ years to meet the growing demand for identifying missing children. We obtained the Children Longitudinal Face (CLF) dataset containing 3,682 face images of 919 children in the age group of $[2, 18]$ years with an average of 4 images per subject collected over an average time span of 4.2 years. Longitudinal performance of three state-of-the-art face recognition systems, COTS-A, FaceNet, and Fused were evaluated. To improve FaceNet's performance on child face images, it was fine-tuned on a training dataset of 3,294 images of 1,119 children (different from CLF dataset). Longitudinal accuracies were evaluated under both verification and open-set identification scenarios. A multilevel statistical model was fit to genuine scores for child face images that included time lapse and gender covariates. Our contributions can be summarized as follows:
\begin{itemize}
\itemsep-1pt 
\item Identification of missing children is viable using current state-of-the-art face matchers, however, improvement in overall face recognition performance of children is much desired. Face verification accuracy for a time lapse of 1 year is high (TAR of 90.18\% at 0.1\% FAR for Fused), but degrades to 73.33\% TARs at 0.1\% FAR after 3 years of elapsed time between enrollment and probe image of a child. We found that the identification performance also decreases over time, however, the rate of degradation in accuracy is small. Detection and Identification Rate (DIR) at a time lapse of 1 year is 79.01\% at 1\% FAR (Rank-3) for Fused. After a 7 year time lapse, DIR drops to 76.42\% for the same FAR and Rank for Fused.
\item We estimate that 80\% of the population in the Children Longitudinal Face dataset can be successfully recognized at 0.1\% FAR by COTS-A, FaceNet, and Fused face matchers for an elapsed time of 2.5, 2, and 2.5 years, respectively.
\item Differences due to gender are matcher-independent. Rates of change in genuine scores for boys are significantly steeper than girls. With higher overall genuine scores, girls in the CLF dataset appear to be easier to recognize than boys.
\end{itemize}

Given the growing concerns about child labor and sex-trafficking, it is essential that we develop and evaluate robust and accurate face recognition systems appropriate to identify missing children. Our longitudinal study is only a small step in this direction. We hope it will stimulate similar studies on a larger collection of children face datasets. A longitudinal study such as ours needs to be conducted periodically to assess current state-of-the-art in age-invariant child face recognition.

{\small
\bibliographystyle{ieeetr}
\bibliography{egbib}
}

\end{document}